\RequirePackage[svgnames]{xcolor}
\documentclass[11pt,letterpaper]{mystyle}

\usepackage[all]{hypcap}
\usepackage[svgnames]{xcolor}
\usepackage[numbers,sort&compress]{natbib}
\usepackage{hyperref}[citecolor=lightblue]

\hypersetup{
    colorlinks = true,
    citecolor = {DodgerBlue},
}
\usepackage{multicol}
\usepackage{setspace}
\usepackage{caption}
\usepackage{ragged2e}
\newcommand{\x}{\mathbf{x}}

\definecolor{blanchedalmond}{rgb}{1.0, 0.92, 0.8}
\definecolor{carmine}{rgb}{0.59, 0.0, 0.09}
\definecolor{lightblue}{rgb}{0.22,0.45,0.70}%

\renewcommand{\mathbf}{\boldsymbol}

\makeatletter
\def\Ddots{\mathinner{\mkern1mu\raise\p@
\vbox{\kern7\p@\hbox{.}}\mkern2mu
\raise4\p@\hbox{.}\mkern2mu\raise7\p@\hbox{.}\mkern1mu}}
\makeatother

\definecolor{amaranth}{rgb}{0.9, 0.17, 0.31}
\definecolor{antiquebrass}{rgb}{0.8, 0.58, 0.46}
\definecolor{antiquefuchsia}{rgb}{0.57, 0.36, 0.51}
\definecolor{chromeyellow}{rgb}{0.31, 0.47, 0.26}

\newcommand{\github}{\raisebox{-1.5pt}{\includegraphics[height=1.05em]{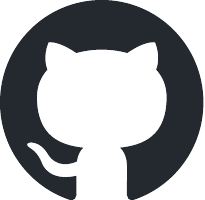}}}

\usepackage{amssymb,mathrsfs,amsmath}

\usepackage[utf8]{inputenc} % allow utf-8 input
\usepackage[T1]{fontenc}    % use 8-bit T1 fonts
\usepackage{url}            % simple URL typesetting
\usepackage{booktabs}       % professional-quality tables
\usepackage{amsfonts}       % blackboard math symbols
\usepackage{nicefrac}       % compact symbols for 1/2, etc.
\usepackage{microtype}      % microtypography
\usepackage{array}
\usepackage{graphicx}
\usepackage{wrapfig}
\usepackage{bxcoloremoji}
\usepackage{pifont}
\usepackage{subcaption}
\usepackage{enumitem}
\usepackage[skip=12pt plus 2pt minus 1pt]{parskip}

\usepackage{booktabs}
\usepackage{multirow}
\usepackage{makecell} % 换行列名
\usepackage{xcolor} % 行高亮
\usepackage{tabularx}

\definecolor{lightgraycustom}{gray}{0.8}
\usepackage{tcolorbox}

\title{MiroThinker: Pushing the Performance Boundaries of Open-Source Research Agents via Model, Context, and Interactive Scaling}

\author{
  MiroMind Team \\
}
\runningtitle{MiroThinker v1.0 Technical Report}

\begin{document}

\begin{abstract}

We present MiroThinker v1.0, an open-source research agent designed to advance tool-augmented reasoning and information-seeking capabilities.
Unlike previous agents that only scale up model size or context length, MiroThinker explores interaction scaling at the model level—systematically training the model to handle deeper and more frequent agent–environment interactions as a third dimension of performance improvement.
Unlike LLM test-time scaling, which operates in isolation and risks degradation with longer reasoning chains, interactive scaling leverages environment feedback and external information acquisition to correct errors and refine trajectories.
Through reinforcement learning, the model achieves efficient interaction scaling: with a 256K context window, it can perform up to 600 tool calls per task, enabling sustained multi-turn reasoning and complex real-world research workflows.
Across four representative benchmarks—GAIA, HLE, BrowseComp, and BrowseComp-ZH—the 72B variant achieves up to 81.9\%, 37.7\%, 47.1\%, and 55.6\% accuracy respectively, surpassing previous open-source agents and approaching commercial counterparts such as GPT-5-high.
Our analysis reveals that MiroThinker benefits from interactive scaling consistently: research performance improves predictably as the model engages in deeper and more frequent agent–environment interactions, demonstrating that interaction depth exhibits scaling behaviors analogous to model size and context length.
These findings establish interaction scaling as a third critical dimension for building next-generation open research agents, complementing model capacity and context windows.

\vspace{2mm}

% \coloremojicode{1F4C5} \textbf{Date}: Nov 13, 2025

\coloremojicode{1F310} \textbf{Online Demo}: \href{https://dr.miromind.ai/}{https://dr.miromind.ai}

\github{} \textbf{Code Repository}: \href{https://github.com/MiroMindAI/MiroThinker}{https://github.com/MiroMindAI/MiroThinker}

\coloremojicode{1F917} \textbf{Model Weights}: \href{https://huggingface.co/miromind-ai/MiroThinker-v1.0-72B}{https://huggingface.co/miromind-ai/MiroThinker-v1.0-72B}

\end{abstract}

\maketitle
\vspace{3mm}
\begin{figure}[h]
\centering
\includegraphics[width=0.99\textwidth]{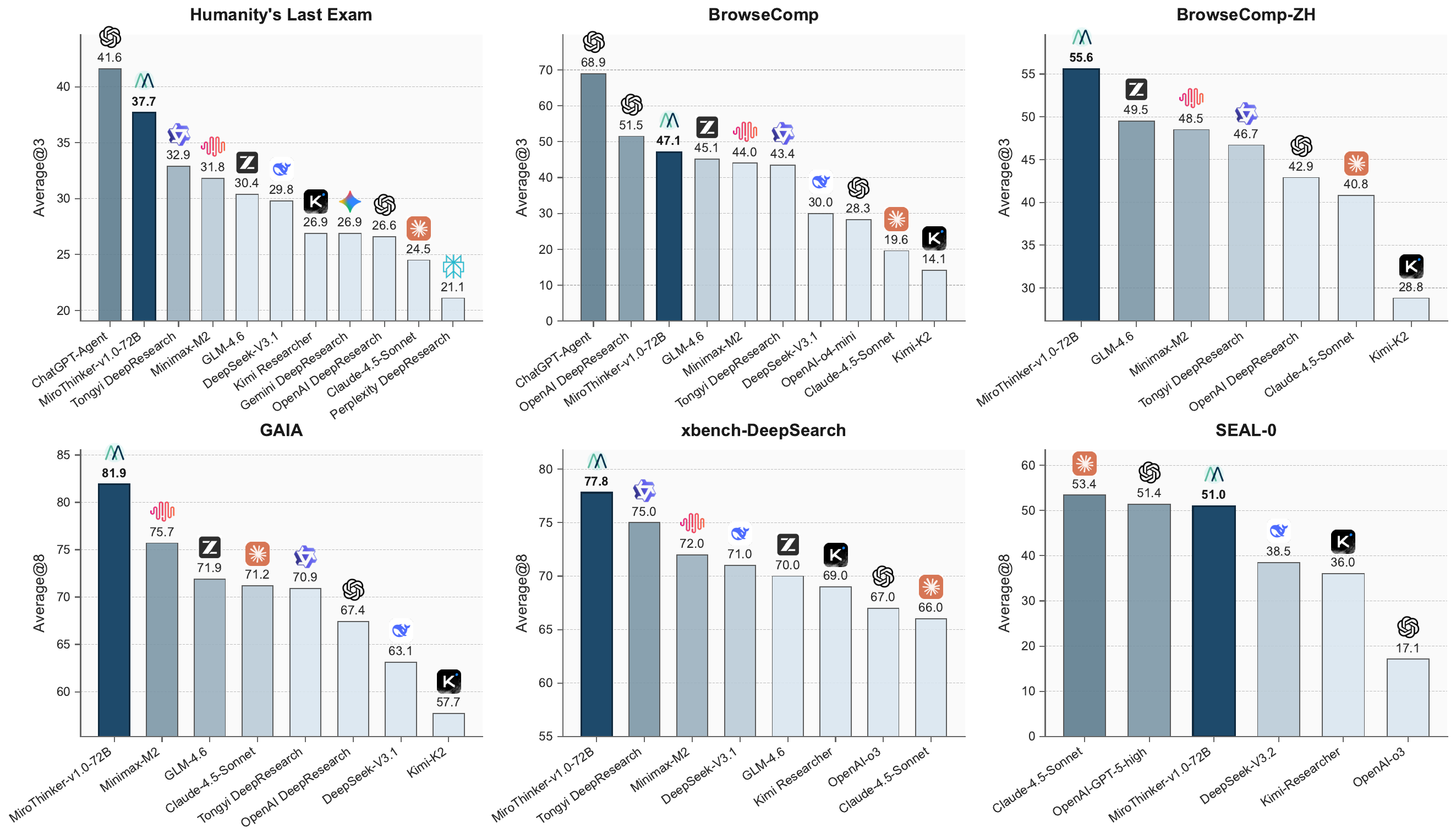}
\caption{Comparison of MiroThinker with state-of-the-art agents and agentic foundation models.}
\label{fig:teaser}
\end{figure}

\section{Introduction}

The rapid evolution of Large Language Models (LLMs) has sparked a paradigm shift in artificial intelligence, from static text generators to dynamic, tool-augmented agents capable of reasoning and interacting with the real world~\citep{openai2025gpt5, team2025kimik2, minimax2025m2, zeng2025glm45, liu2024deepseekv3, yang2025qwen3, anthropic2025claude4_5}. 
Within this emerging paradigm, research capability has become a new frontier of intelligence. Achieving research-level reasoning requires more than linguistic fluency; it demands the ability to formulate hypotheses, retrieve and verify evidence, and synthesize insights across diverse information sources. 
Proprietary systems such as ChatGPT Agent~\citep{openai2025chatgpt_agent} and Claude Research~\citep{anthropic2025claude_research} exemplify the potential of such capabilities, demonstrating near-human proficiency in literature review, comparative analysis, and reasoning-driven knowledge discovery. However, these systems remain closed, constraining transparency, reproducibility, and community-driven innovation.

To narrow the gap with proprietary systems, the open-source community has made remarkable progress in recent years.
At the foundation model level, open-weight LLMs are increasingly trained with built-in agentic skills such as search, browsing, and coding~\citep{liu2024deepseekv3, team2025kimik2, zeng2025glm45, minimax2025m2, team2025longcat}. Yet these projects typically release only model weights without providing the full suite of tools or agentic frameworks necessary for end-to-end research reasoning.
Meanwhile, a parallel line of open-source work~\citep{team2025tongyi, li2025webthinker, li2025websailor, tao2025webshaper, fang2025cognitive, li2025afm, tang2025beyond, wu2025webdancer, zheng2025deepresearcher, song2025r1searcher, liu2025webexplorer, zhang2025infoagent} focuses on developing specialized research agent models, along with their corresponding toolchains and frameworks. However, these models are often relatively small in scale and constrained in context length and interaction depth, leaving a clear performance gap compared with leading commercial research agents~\citep{openai2025chatgpt_agent, anthropic2025claude_research, moonshot2025kimi_researcher, openai2025deepresearch}.

In response to these challenges, we introduce \texttt{MiroThinker v1.0}, an open-source, high-performance research agent model that pushes the performance boundaries of open-source systems along three key dimensions: \emph{model size}, \emph{context length}, and \emph{interaction depth}.
Leveraging these capabilities, the model engages in iterative cycles of reasoning and tool use, enabling it to decompose complex research problems, retrieve and integrate real-time information from the Internet, synthesize multi-source evidence, and generate transparent, well-grounded conclusions.
\textbf{To sustain such deep reasoning processes, the model is equipped with a 256K context window, supporting up to 600 tool calls per task, a significant leap from the previous open-source models of fewer than 100.}
To support diverse computational budgets, we release MiroThinker v1.0 in 8B, 30B, and 72B variants, together with a comprehensive suite of tools.

As shown in Figure~\ref{fig:teaser}, empirical evaluations demonstrate that MiroThinker v1.0, equipped with a simple ReAct agent, achieves state-of-the-art performance among open-source research agents, approaching the results of leading commercial systems~\cite{moonshot2025kimi_researcher, anthropic2025claude_research, openai2025chatgpt_agent, openai2025deepresearch, openai2025gpt5}.
Specifically, on the BrowseComp~\citep{wei2025browsecomp} benchmark, MiroThinker-v1.0-72B achieves 47.1\% accuracy, surpassing MiniMax-M2~\cite{minimax2025m2} by 2.0 points.
A similar pattern is observed on the BrowseComp-ZH~\citep{zhou2025browsecomp_zh} benchmark, with a 6.1-point improvement over GLM-4.6~\cite{zeng2025glm45}, underscoring the model's robust multilingual reasoning ability.
Further, on Humanity's Last Exam (HLE)~\citep{phan2025hle}, MiroThinker-v1.0-72B continues to outperform its peers, achieving a score of 37.7\%, 4.8 points higher than Tongyi-DeepResearch~\cite{team2025tongyi}.
On the GAIA-Text-Only benchmark~\citep{mialon2023gaia}, MiroThinker-v1.0-72B attains a score of 81.9\%, surpassing its strongest open-source counterpart MiniMax-M2~\cite{minimax2025m2} (75.7\%) by 6.2 points.
Overall, across diverse benchmarks, MiroThinker v1.0 maintains a consistent and substantial advantage, demonstrating stronger reasoning capability, long-context comprehension, and deep tool-use proficiency than existing open-source research agents.
\section{Related Works}

\paragraph{Agent Foundation Models} 
Recent research has increasingly emphasized enhancing Large Language Models (LLMs) with agentic capabilities—the ability to plan, reason, and act autonomously in complex environments~\citep{openai2025gpt5, team2025kimik2, liu2024deepseekv3, minimax2025m2, zeng2025glm45, yang2025qwen3, team2025longcat, anthropic2025claude4_5}. Building on this trend, Agent Foundation Models (AFMs) have emerged as a new paradigm of foundation models that, beyond learning general language understanding, explicitly incorporate agent-oriented abilities, such as decision-making, tool use, and interaction with external environments, during their base model training. Current efforts particularly focus on code and search agents, aiming to enhance capabilities in tool-based problem solving, retrieval-augmented reasoning, and autonomous task execution. Representative AFMs such as GPT-5~\citep{openai2025gpt5}, Claude-4.5~\citep{anthropic2025claude4_5}, Grok-3~\citep{xai2025grok3}, Kimi K2~\citep{team2025kimik2}, MiniMax M2~\citep{minimax2025m2}, GLM-4.6~\citep{zeng2025glm45}, and DeepSeek-V3.1~\citep{liu2024deepseekv3} have demonstrated promising progress in these domains, underscoring a shift toward more specialized and practical forms of agentic intelligence.

\paragraph{Deep Research Models}  
Continuing this progression, deep research models have been introduced as specialized LLM-based agents for complex multi-hop reasoning and long-context, retrieval-intensive tasks. These models integrate dynamic information-seeking and iterative planning into their workflows, enabling autonomous acquisition and synthesis of knowledge into comprehensive answers. Major AI labs have accordingly developed proprietary deep research systems: OpenAI Deep Research~\citep{openai2025deepresearch}, Claude Research~\citep{anthropic2025claude_research}, Kimi-Researcher~\citep{moonshot2025kimi_researcher}, Grok DeepSearch~\citep{xai2025grok3}, \emph{etc.}, all extend LLMs with agentic tool use and long-horizon reasoning tailored for in-depth research. 
Meanwhile, the open-source community has introduced many deep research models. For example, WebThinker~\citep{li2025webthinker}, WebSailor~\citep{li2025websailor}, WebShaper~\citep{tao2025webshaper}, and Tongyi DeepResearch~\citep{team2025tongyi} leverage LLMs with iterative web-browsing workflows, whereas Cognitive Kernel-Pro~\citep{fang2025cognitive}, AFM~\citep{li2025afm}, WebDancer~\citep{wu2025webdancer}, and DeepMiner~\citep{tang2025beyond}, \emph{etc.}, explore novel training algorithms and dynamic memory mechanisms to push the boundaries of research capabilities. Collectively, these developments underscore a broader shift toward LLMs that serve as specialized research assistants, combining advanced reasoning with real-time information retrieval to tackle open-ended knowledge-intensive tasks.

\section{Agentic Workflow}

\subsection{Formulation}

MiroThinker v1.0 model is developed under the ReAct paradigm~\citep{yao2022react} in a single-agent setting. 
Given a query $q$, the model alternates between reasoning, tool invocation, and observation in an iterative loop until termination. 
At step $t$, the agent maintains a trajectory
\begin{equation}
H_t = \{(T_1, A_1, O_1), \ldots, (T_{t-1}, A_{t-1}, O_{t-1})\},
\end{equation}
where $T_i$, $A_i$, and $O_i$ denote the thought, action, and observation, respectively. 
The thinking model $f_\theta$ generates an internal thought context $T_t = f_\theta(q, H_t)$, 
followed by an action policy
\begin{equation}
A_t = \pi_\theta(H_t, T_t),
\end{equation}
which takes as input the trajectory history $H_t$ and the current thought $T_t$, and outputs a structured tool invocation that specifies which external tool to use and how to query it. 
The environment executes the invocation and returns a tool response $O_t = \text{Tool}(A_t)$, 
which is appended to form
\begin{equation}
H_{t+1} = H_t \cup \{(T_t, A_t, O_t)\}.
\end{equation}
This reasoning–acting–observing loop continues until the model outputs no further action ($A_t = \emptyset$), 
upon which a summary phase produces the final answer $y = g_\theta(H_t)$. 
This iterative workflow enables dynamic reasoning grounded in external evidence, yielding interpretable and adaptive decision-making compared with static single-pass LLMs.

\begin{figure}[t]
\centering
\includegraphics[width=0.93\textwidth]{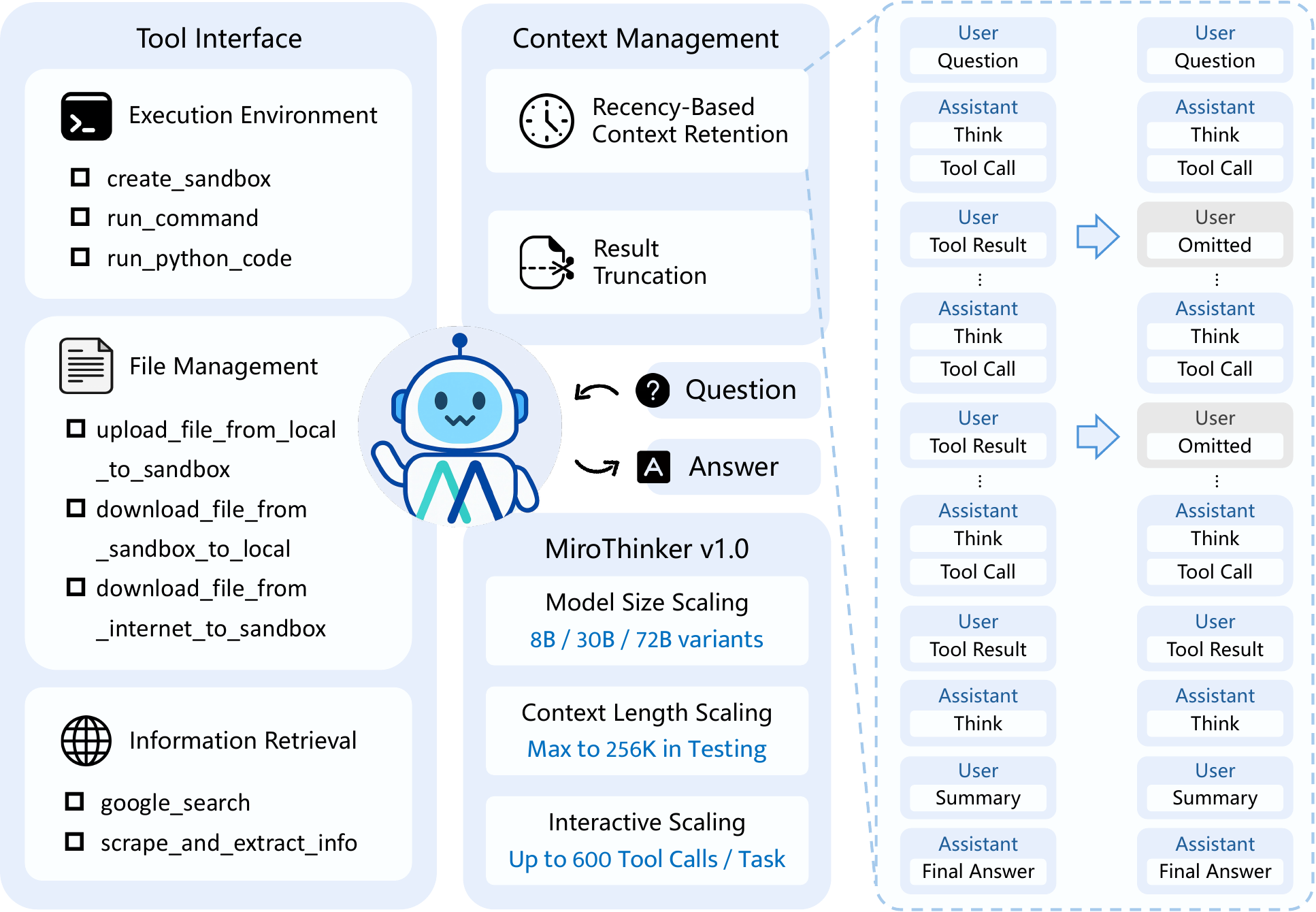}
\caption{Overview of the MiroThinker v1.0 agent architecture. The framework integrates a structured tool interface, \emph{i.e.,} execution environment, file management, and information retrieval, with a simple recency-aware context management to support interactive scaling. On the right, an agentic trajectory example illustrates the recency-based context retention mechanism, where tool outputs from earlier turns are omitted to maintain context efficiency.}
\label{fig:overview}
\end{figure}

\subsection{Tool Interface}

To enable interaction with external environments, the model is equipped with a modular tool interface that exposes a set of tools. 
Each tool encapsulates a specific capability (\emph{e.g.}, code execution, file handling, or web retrieval), allowing the model to act beyond pure text generation.

\paragraph{Execution Environment} 
We employ a Linux sandbox that provides an isolated runtime for command and code execution. 
The agent can create a sandbox instance via \texttt{create\_sandbox} and subsequently execute shell commands (\texttt{run\_command}) or Python code (\texttt{run\_python\_code}) within it. 
This design ensures safe and flexible interaction with system-level resources.

\paragraph{File Management}
To facilitate data flow between the sandbox and the external world, we implement file upload and download utilities. 
Specifically, \texttt{upload\_file\_from\_local\_to\_sandbox} and \texttt{download\allowbreak\_file\allowbreak\_from\allowbreak\_sandbox\allowbreak\_to\allowbreak\_local} support bidirectional file transfer, 
while \texttt{download\allowbreak\_file\allowbreak\_from\allowbreak\_internet\allowbreak\_to\allowbreak\_sandbox} enables direct retrieval of remote assets from a given URL.

\paragraph{Information Retrieval}
For knowledge-intensive reasoning, the agent is equipped with two retrieval tools: 
a Google-based web search tool (\texttt{google\_search}) that returns structured search results, 
and a web-scraping tool (\texttt{scrape\_and\_extract\_info}) for conditional information extraction from target URLs. 
Unlike naive webpage scraping, this tool internally leverages a lightweight LLM (\emph{e.g.}, Qwen3-14B~\citep{yang2025qwen3}) to extract task-relevant information specified by the agent at call time. 
This mechanism serves as an efficient form of context management, allowing the tool to condense lengthy web or document content into focused textual evidence for subsequent reasoning.
Note, to prevent potential information leakage (\emph{e.g.}, searching benchmark answers from HuggingFace), access to HuggingFace has been explicitly disabled in these tools.

\subsection{Context Management}

To ensure efficient use of the model’s context window, we apply two strategies to manage tool responses. These strategies enable our model to perform up to 600 tool calls within a 256K context window.

\paragraph{Recency-Based Context Retention}
In the standard ReAct paradigm~\cite{yao2022react}, all tool outputs are retained in the message history, often leading to inefficient context utilization. 
Empirically, we observe that the model’s subsequent actions depend primarily on recent observations rather than distant ones. 
To leverage this recency bias and improve contextual efficiency, we retain only the most recent tool responses while preserving the complete sequence of thoughts and actions.
% To improve context efficiency, we prune early tool responses while retaining the full sequence of thoughts and actions.
Given a retention budget $K \in \mathbb{N}$ for preserved tool responses, define the index set of the most recent responses at step $t$ as
\begin{equation}
S_t(K) = \{\, i \in \{1,\dots,t-1\} \mid i \ge t-K \,\}.
\end{equation}
We construct a recency-filtered history $\widehat{H}_t$ by masking tool responses outside $S_t(K)$:
\begin{equation}
\widehat{H}_t
= \big\{ \big(T_i,\; A_i,\; \widehat{O}_i \big) \big\}_{i=1}^{t-1},
\quad
\widehat{O}_i \triangleq 
\begin{cases}
O_i, & i \in S_t(K),\\
\varnothing, & \text{otherwise},
\end{cases}
\end{equation}
where $\varnothing$ denotes that the early tool response is omitted from the context. 
Subsequent inference is performed on the recency-filtered history:
\begin{equation}
T_t = f_{\theta}(q, \widehat{H}_t), 
\qquad
A_t = \pi_{\theta}(\widehat{H}_t, T_t),
\end{equation}
and upon receiving the new tool response $O_t = \mathrm{Tool}(A_t)$, we update via
\begin{equation}
H_{t+1} = \big\{ (T_1, A_1, O_1), \dots, (T_t, A_t, O_t) \big\},
\qquad
\widehat{H}_{t+1} = \mathrm{Retain}_K(H_{t+1}),
\end{equation}
where $\mathrm{Retain}_K(\cdot)$ applies the masking rule above with budget $K$.
This recency-based retention strategy preserves the reasoning and action trace while focusing the model’s attention on the most contextually relevant observations, thereby freeing additional context for extended reasoning and deeper tool-use trajectories.
We found that such a simple context management strategy serves as a strong baseline; it does not lead to degradation in performance and allows the model to have more context space for interactive scaling.

\paragraph{Result Truncation}
Certain tools, such as \texttt{run\_command} and \texttt{run\_python\_code}, may occasionally produce excessively long outputs that could easily overflow the model’s context. To mitigate this, we truncate tool responses that exceed a predefined length limit and append the tag ``[Result truncated]'' at the end to indicate that the content has been shortened.
\section{Data Construction}

\begin{figure}[t]
\centering
\includegraphics[width=\textwidth]{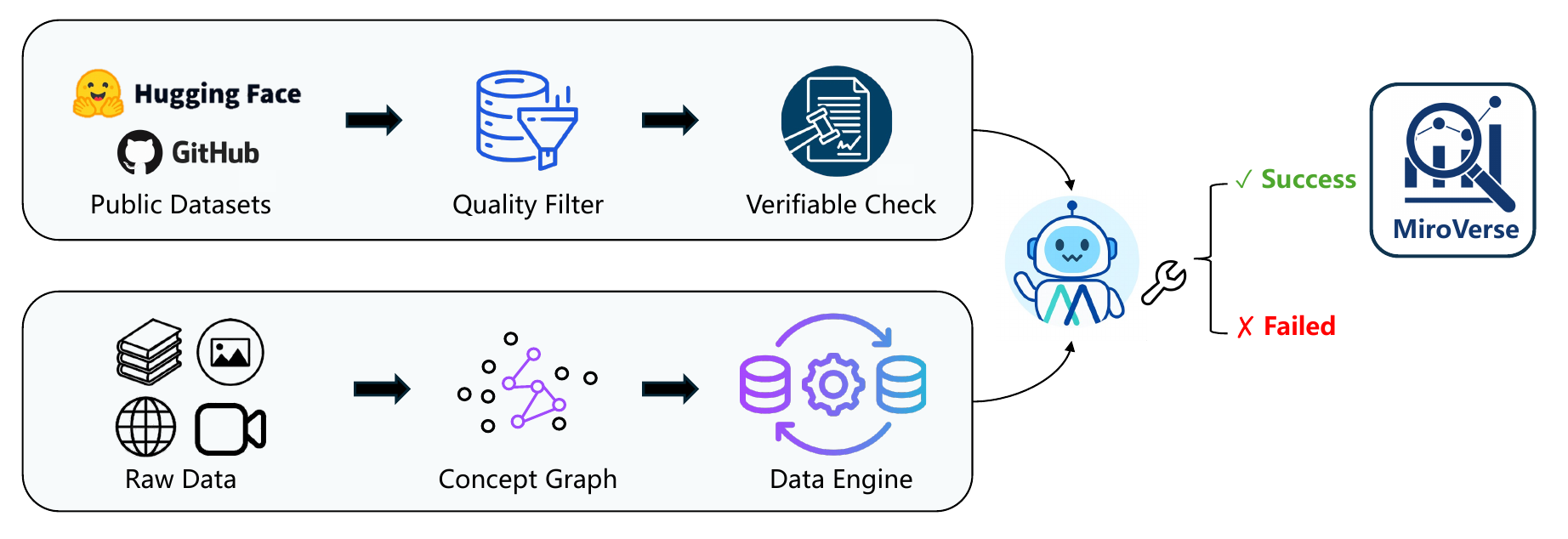}
\caption{Overview of the data construction pipeline. Public datasets from platforms such as HuggingFace and GitHub are filtered and verified, while raw internet data are processed through knowledge graph generation and a data engine. The resulting QA pairs from both sources are then converted into agentic trajectories, forming the complete MiroVerse v1.0 dataset used for training MiroThinker v1.0.}
\label{fig:intro}
\end{figure}

To effectively train our MiroThinker, we construct a large-scale synthetic dataset comprising two main components: (1) multi-document QA (MultiDocQA) synthesis, and (2) agentic trajectory synthesis. These two stages jointly enable the model to acquire both factual grounding and reasoning capabilities.

\subsection{MultiDocQA Synthesis}

We design a question–answer synthesis pipeline that transforms interlinked web documents into complex, multi-hop QA pairs. The overall process is structured into multiple stages, as described below.

\paragraph{Document Corpus Construction} 
We construct the document corpus from diverse, highly interlinked sources such as Wikipedia, Common Crawl, and curated web repositories, chosen for their rich hyperlink structures and factual reliability. 
During preprocessing, we clean textual content while preserving hyperlinks, which form the foundation for constructing multi-document reasoning chains.
Documents are then categorized into broad knowledge domains via a hybrid method combining metadata extraction and topic modeling, enabling category-aware sampling in later stages.

\paragraph{Document Sampling and Graph Construction} 
We begin by sampling document nodes from our corpus while maintaining balanced representation across different categories. 
This category-balanced sampling ensures comprehensive coverage across diverse knowledge domains and prevents bias toward over-represented topics. 
For each sampled seed document, we construct a knowledge graph by following its internal hyperlinks. 
Specifically, we randomly select one internal link from each document and recursively repeat this process multiple times to build a connected subgraph of related documents.

\paragraph{Document Consolidation} 
After constructing the document graph, we convert each document into markdown format and perform link pruning. 
We remove all hyperlinks that point to documents outside our selected subgraph, ensuring that the consolidated article maintains coherent references only within the current context. 
These preprocessed documents are then concatenated into a single comprehensive article that spans multiple related topics while maintaining logical flow through the preserved internal references.

\paragraph{Fact Extraction} 
For each document in the constructed graph, we identify key factual statements that connect back to the central theme established by the seed document. 
This targeted extraction process ensures that the facts we collect form a coherent knowledge network rather than isolated pieces of information. 
We prioritize statements that inherently require cross-document reasoning to discover or verify, thereby establishing a foundation for questions that cannot be answered from any single source alone. 
These extracted facts collectively represent the multi-hop knowledge required to answer complex questions in our dataset.

\paragraph{Constraint Obfuscation}
To create challenging reasoning scenarios, we systematically obfuscate the extracted facts by transforming them into indirect constraints that require deeper reasoning to resolve. 
Our obfuscation strategies operate differently depending on the type of information: temporal and spatial details are generalized to broader categories (\emph{e.g.}, March 15, 2023 → in the spring of the 2020s; Paris → a European capital), while other entities and concepts are expressed through referential indirection using related properties or contextual descriptions. 
This transformation compels models to perform multi-step associative reasoning, integrating knowledge across multiple sources rather than relying on direct fact retrieval.

\paragraph{Question Generation} 
Finally, we prompt a large language model to synthesize questions by selecting and combining multiple obfuscated constraints from our fact pool. 
The LLM is instructed to generate questions that span diverse domains and require integration of information across different documents in the knowledge graph. 
This ensures that the resulting questions demand genuine multi-hop reasoning capabilities and cannot be answered through simple pattern matching or single-document retrieval.

\subsection{Agentic Trajectory Synthesis}
\label{sec:trace_synth}
To generate high-quality and diverse agentic trajectory data, we design a multi-layered synthesis framework that integrates multiple agent paradigms, tool invocation mechanisms, and state-of-the-art LLMs.

\paragraph{Agent Paradigms}
We employ two complementary agent paradigms for trajectory synthesis:

(1) ReAct Single-Agent~\citep{yao2022react}. This paradigm addresses complex tasks through iterative ``think–act–observe'' cycles. The agent first analyzes the current state and reasons about the next action, then executes tool calls, and finally updates its internal understanding based on observations. This approach is particularly effective for tasks that require multi-step reasoning and adaptive decision-making.

(2) MiroFlow Multi-Agent~\citep{2025miroflow}. This framework coordinates multiple specialized agents to manage complex workflows. Each agent handles distinct subtasks or domains, communicating through structured protocols. This paradigm produces sophisticated collaborative trajectories that exhibit division of labor, coordination, and emergent collective reasoning.

\paragraph{Tool Invocation Mechanisms}
To enhance the diversity of synthesized trajectories, we adopt a hybrid approach that combines two complementary tool invocation methods:

(1) Function Calling. A traditional, structured tool invocation approach in which agents interact with external tools through predefined function interfaces. 
This method provides clear input-output specifications and is suitable for standardized tool usage scenarios.

(2) Model Context Protocol (MCP). A more flexible tool invocation protocol that enables agents to interact with tools more naturally through context negotiation. MCP supports more complex tool composition and dynamic tool discovery, making synthesized trajectories closer to real human-machine interaction patterns.

\paragraph{Diverse Data Synthesis}
We employ multiple leading LLMs to drive the trajectory synthesis process, including GPT-OSS~\citep{agarwal2025gptoss}, DeepSeek-V3.1~\citep{liu2024deepseekv3}, and other state-of-the-art models. 
By employing diverse models to generate trajectories, we obtain training data with diverse styles, mitigating single-model biases and ensuring both richness and coverage.

\subsection{Open-Source Data Collection}

We supplement our synthesized data with diverse open-source QA datasets to broaden coverage and enhance reasoning diversity. The incorporated datasets include MuSiQue~\citep{trivedi2022musique}, HotpotQA~\citep{yang2018hotpotqa}, WebWalkerQA-Silver~\citep{wu2025webwalker}, MegaScience~\citep{fan2025megascience}, TaskCraft~\citep{shi2025taskcraft}, QA-Expert-Multi-Hop-V1.0~\citep{khaimaitien2023qaexpert}, OneGen-TrainDataset-MultiHopQA~\citep{zjunlp2024onegen}, 2WikiMultihopQA~\citep{ho2020twowikimultihop}, WikiTables~\citep{kweon2023wikitable}, WebShaper~\citep{tao2025webshaper}, WebDancer~\citep{wu2025webdancer}, and Toucan-1.5M~\citep{xu2025toucan}.
We retain only the QA pairs from these datasets and convert them into agentic trajectories through the synthesis pipeline described in Section~\ref{sec:trace_synth}. 
To preserve general conversational abilities, we also include post-training corpora such as AM-Thinking-v1-Distilled~\citep{tian2025amthinking} and Nemotron-Post-Training-Dataset~\citep{basant2025nemotron}, providing broad coverage of reasoning styles and dialogue forms.

\section{Training Pipeline}

Based on the open-source Qwen2.5 and Qwen3 models~\cite{yang2025qwen3}, MiroThinker is trained under a three-stage pipeline:
(1) Supervised fine-tuning to establish fundamental agentic behaviors;
(2) Preference optimization to align decision-making with task objectives;
(3) Reinforcement learning to drive creative exploration and generalization in real-world environments.

\subsection{Agentic Supervised Fine-tuning}

The first stage performs supervised fine-tuning (SFT) to endow MiroThinker with agentic behaviors.
The model learns to mimic expert trajectories that involve multi-hop reasoning and tool use. 

\paragraph{Data Construction}
We construct a large-scale SFT dataset $\mathcal{D}_{\text{SFT}} = \{(x_i, H_i)\}_{i=1}^N$, where each instance pairs a task instruction $x_i$ with an expert trajectory 
$H_i = \{(T_{i,t}, A_{i,t}, O_{i,t})\}_{t=1}^{T_i}$ composed of thought–action–observation triplets. 
During this process, we observe that even when synthesized using leading LLMs, the raw trajectories often contain substantial noise, such as intra-response repetition, cross-response duplication, and invalid tool invocations (\emph{e.g.}, incorrect tool names or malformed arguments). 
To mitigate these issues, we apply rigorous filtering and data-repair procedures to ensure the consistency and reliability of the final SFT corpus.

\paragraph{Training Objective}
Each trajectory is treated as a multi-turn dialogue between a \emph{user} and an \emph{assistant}. 
The user provides the initial task instruction $x$ and subsequently the tool observations $O_t$, while the assistant produces the reasoning thoughts $T_t$ and tool invocations $A_t$. 
During training, tool execution is not actually performed; the observations are pre-recorded and serve as contextual inputs. 
Given $(x, H) \sim \mathcal{D}_{\text{SFT}}$, the model is trained to predict the expert’s thought and action sequences:
\begin{equation}
\mathcal{L}_{\text{SFT}}(\theta)
= -\mathbb{E}_{(x,H)} \left[ \sum_{t=1}^{T_H} \log \pi_\theta(T_t, A_t \mid x, H_{<t}) \right].
\end{equation}
This formulation aligns the agent’s imitation learning with standard dialogue-style SFT, 
where tool responses are treated as user turns and the assistant learns to generate the next reasoning or tool call accordingly.

\subsection{Agentic Preference Optimization}

The second stage refines decision-making through Direct Preference Optimization (DPO)~\citep{rafael2023dpo}, using preference data synthesized from the SFT model.

\paragraph{Data Collection}
We construct a pairwise preference dataset
\begin{equation}
\mathcal{D}_{\text{PO}} = \{(x_i, H_i^{+}, H_i^{-})\}_{i=1}^{M},
\end{equation}
where each task instruction $x_i$ is associated with a preferred trajectory $H_i^{+}$ and a dispreferred trajectory $H_i^{-}$. 
Each trajectory represents a full multi-step interaction in thought–action–observation space. Preferences are determined based on the following steps:

(1) {Criterion: Correct \emph{vs.} Incorrect Without Enforced Patterns}. We construct preference pairs based primarily on the correctness of the final answer. Unlike approaches that rely on handcrafted heuristics or fixed agentic patterns, \emph{e.g.}, predefined planning length, step counts, or reasoning structures, we find that such constraints introduce systematic biases and hinder scalability across diverse tasks and domains. Therefore, we avoid imposing rigid structural formats and instead rely on the correctness of the answer for ranking preferences.

(2) {Quality Control: Ensuring Trace Completeness}. We further apply strict filtering to ensure the quality and faithfulness of both chosen and rejected trajectories. For a chosen sample, the reasoning trace must be coherent, include an explicit planning process, and yield a clear and correct final answer. For a rejected sample, we similarly require that the trajectory produce a valid final answer. In addition, we apply further filtering to remove surface-level issues such as repetition, truncation, or malformed structures, ensuring that only high-quality trajectories are retained.

\paragraph{Training Objective}
We refine the SFT model using DPO augmented with an auxiliary SFT loss on preferred trajectories~\cite{liu2024rpo, wang2024mpo}, to enhance stability and preserve behavioral consistency. Given a task instruction $x$ and a preference pair $(H^+, H^-)$, the DPO objective encourages the model to assign a higher likelihood to the preferred trajectory while remaining close to the reference SFT model:
\begin{equation}
\mathcal{L}_{\text{DPO}}(x,H^+,H^-)
= - \log \sigma \!\left(
\beta \!\left[
(\log \pi_\theta(H^+|x) - \log \pi_\theta(H^-|x))
- (\log \pi_{\text{ref}}(H^+|x) - \log \pi_{\text{ref}}(H^-|x))
\right]
\right),
\end{equation}
where $\pi_{\text{ref}}$ is the frozen reference model and $\beta$ controls deviation from it.
The full objective combines DPO with the SFT loss defined above, applied to preferred samples:
\begin{equation}
\mathcal{L}_{\text{PO}}(\theta)
= \mathbb{E}_{(x,H^+,H^-)} [L_{\text{DPO}}(x,H^+,H^-)]
+ \lambda \, \mathcal{L}_{\text{SFT}}^{(+)}(\theta),
\end{equation}
where $\mathcal{L}_{\text{SFT}}^{(+)}$ denotes the SFT loss on preferred trajectories, and $\lambda$ controls its weight.

\subsection{Agentic Reinforcement Learning}

\begin{figure}[t]
  \centering
  \begin{subfigure}{0.45\linewidth}
    \includegraphics[width=\linewidth]{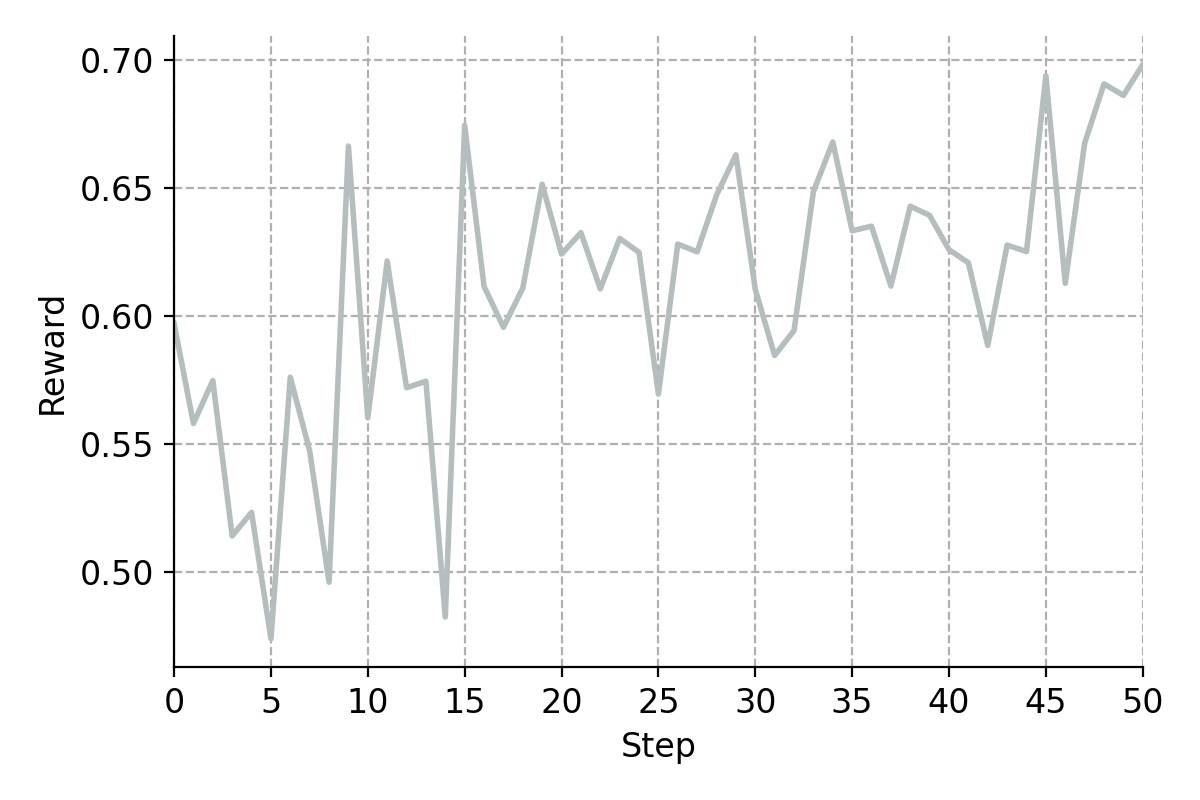}
    \caption{Training reward across training steps.}
    \label{fig:grpo_reward}
  \end{subfigure}
  % \hfill
  \hspace{2em}
  \begin{subfigure}{0.45\linewidth}
    \includegraphics[width=\linewidth]{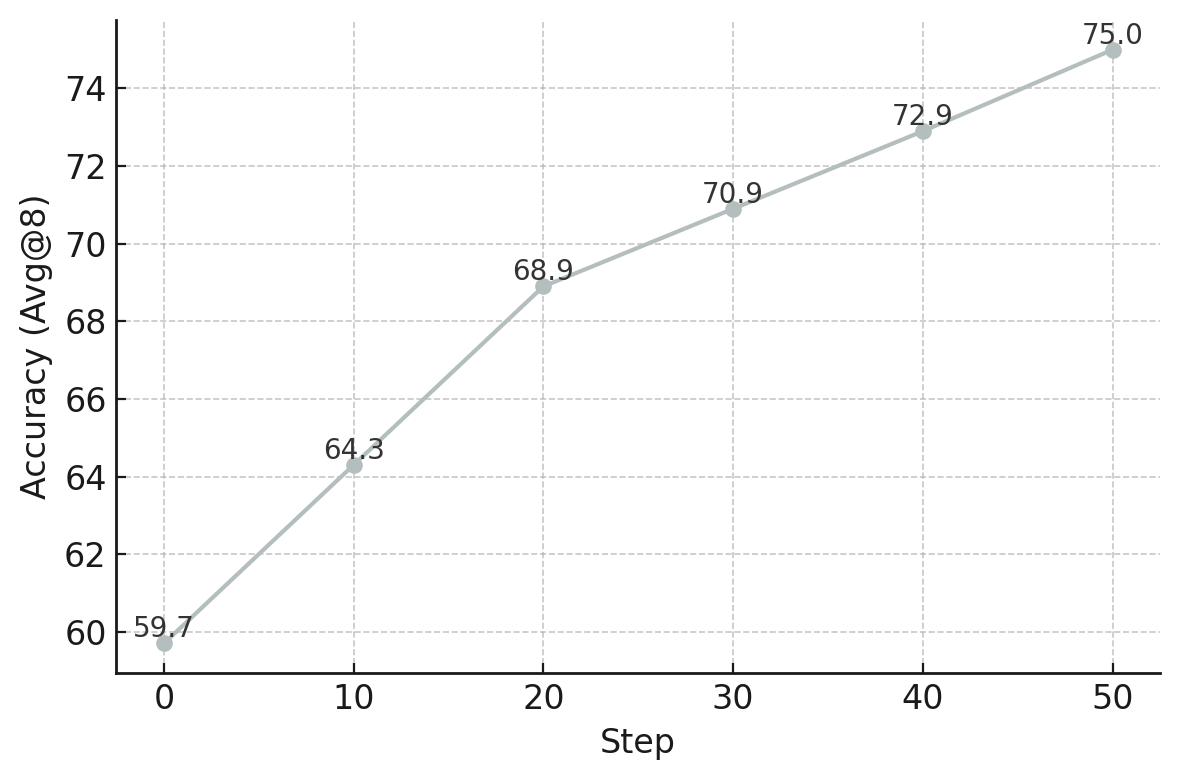}
    \caption{Val acc on GAIA-Text-103 across training steps.}
    \label{fig:grpo_val}
  \end{subfigure}

\label{fig:grpo}
\caption{Training dynamics of MiroThinker-v1.0-30B for GRPO Agentic RL.
Since the RL environment is not exactly the same as the final evaluation environment, there will be slight differences in performance.
}

\end{figure}

The final stage leverages reinforcement learning to enable the agent to discover creative solutions and adapt to diverse real-world environments through direct interaction and exploration. 
We employ Group Relative Policy Optimization (GRPO)~\cite{shao2024deepseekmath} with fully online policy training by using rollout trajectories to update the policy model exactly once.

\paragraph{Environment Setup} 
We construct a suite of scalable environments capable of supporting thousands of concurrent agentic rollouts, encompassing real-time multi-source search, web scraping and summarization, Python code execution, and Linux VM manipulation.
We also built a robust and knowledgeable LLM grading system for verifying noisy agent predictions against ground-truth answers in low latency.

\paragraph{Streaming Rollout Acceleration}
Unlike single-turn RL tasks like math or reasoning, agentic RL requires a multi-round back-and-force between LLMs and environments, making the completion time of different trajectories heavily long-tailed.
The tailness problem is further amplified as our MiroThinkers are capable of interacting with the environments for hundreds of rounds.
We implement a rollout mechanism where each agent worker receives prompts from a task queue in a streaming manner until enough completed trajectories are collected for this batch. 
All unfinished tasks are pushed back to the task queue for the next iteration.

\paragraph{Reward Design}
Our reward function $R(x, H)$ for a trajectory $ H = \{(T_t, A_t, O_t)\}_{t=1}^{T_H} $ given question $x$ combines multiple components:
\begin{equation}
R(x, H) = \alpha_c R_\text{correct}(H)
- \alpha_f R_\text{format}(H),
\end{equation}
where $R_\text{correct}$ measures solution correctness and $R_\text{format}$ penalizes the model for failing to follow format instruction. 
The coefficients $\{\alpha_c, \alpha_f\}$ are tuned to balance persistent exploration of new solutions with the ability of instruction following.

\paragraph{Trajectory Curation}
To ensure RL learning signal quality, we implement a comprehensive trajectory filtering pipeline that removes both noisy correct trajectories and trivially incorrect ones.
For correct trajectories, we filter out samples exhibiting pathological behaviors such as consecutive API call failures (\emph{e.g.}, more than 5 consecutive network exceptions), redundant retries on identical actions, or excessive environment timeout errors, as these patterns do not reflect genuine problem-solving strategies but rather environmental instabilities.
For incorrect trajectories, we discard samples failing due to trivial formatting issues (\emph{e.g.}, missing required answer format) or exhibiting degenerate behaviors like action repetition loops or premature termination without meaningful exploration.

\paragraph{Training Objective}
GRPO optimizes the policy by sampling multiple trajectories per prompt and computing advantages relative to the group mean. For each prompt $x$, we sample a group of $G$ trajectories $\{H_1, \ldots, H_G\}, \quad H_i = \{(T_{i,t}, A_{i,t}, O_{i,t})\}_{t=1}^{T_i} $
from the current policy $\pi_\theta$. The advantage for trajectory $H_i$ is computed as:
\begin{equation}
\hat{A}_i = R(x, H_i) - \frac{1}{G}\sum_{j=1}^G R(x, H_j).
\end{equation}
The GRPO objective maximizes the expected advantage while maintaining proximity to the reference policy:
\begin{equation}
\mathcal{L}_\text{GRPO}(\theta) =
\mathbb{E}_{x \sim \mathcal{D}}
\mathbb{E}_{H \sim \pi_\theta(\cdot \mid x)} \left[
\hat{A}(x, H) \cdot \log \pi_\theta(H \mid x)
- \beta_\text{KL} \cdot D_\text{KL}\big(\pi_\theta(\cdot \mid x) \,\|\, \pi_\text{ref}(\cdot \mid x)\big)
\right],
\end{equation}
where $\pi_\text{ref}$ is the reference policy (typically the preference optimization checkpoint), and $\beta_\text{KL}$ controls the strength of the KL penalty.

\begin{table}[t]
\centering
\footnotesize
\definecolor{rowhl}{RGB}{225,237,237}
\caption{Performance comparison across various agent benchmarks. Performance of other Agent Foundation models (AFMs) is collected from Tongyi DeepResearch~\cite{team2025tongyi} and model cards of DeepSeek-V3.1, DeepSeek-V3.2, MiniMax-M2, and Kimi-K2. 
To minimize the impact of randomness from agent-environment interactions on benchmark performance evaluation, we report the average performance and error bars for each benchmark. 
We use avg@3 for Humanity’s Last Exam, BrowseComp, BrowseComp-ZH, WebWalkerQA, and FRAMES, and avg@8 for GAIA, xbench-DeepSearch, and SEAL-0.}
\label{tab:benchmarks-fast}
\renewcommand\arraystretch{1.1}
\setlength{\tabcolsep}{4pt}
\begin{tabularx}{\linewidth}{@{}lcccccccc@{}}
\toprule
\textbf{Benchmarks} &
\makecell{\textbf{Humanity's}\\\textbf{Last Exam}} &
\makecell{\textbf{Browse}\\\textbf{Comp}} &
\makecell{\textbf{Browse}\\\textbf{Comp-ZH}} &
\textbf{GAIA} &
\makecell{\textbf{xbench}\\\textbf{DeepSearch}} &
\makecell{\textbf{WebWalker}\\\textbf{QA}} &
\textbf{FRAMES} &
\textbf{SEAL-0} \\
\midrule
\multicolumn{8}{l}{\textit{Foundation Models with Tools}} \\
\midrule
% Results of GLM-4.5 from: https://arxiv.org/pdf/2510.24701
% GLM-4.5~\cite{zeng2025glm45}                       & 21.2 & 26.4 & 37.5 & 66.0 & 70.0 & 65.6 & 78.9 &  --  \\
% Results of GLM-4.6 from: https://huggingface.co/MiniMaxAI/MiniMax-M2
GLM-4.6~\cite{zeng2025glm45}                       & 30.4 & 45.1 & 49.5 & 71.9 & 70.0 & --   & --   &  --  \\
% Results of MiniMax-M2 from: https://huggingface.co/MiniMaxAI/MiniMax-M2
Minimax-M2~\cite{minimax2025m2}                    & 31.8 & 44.0 & 48.5 & 75.7 & 72.0 & --   & --   &  --  \\
% Results of DeepSeek-V3.1 from: https://arxiv.org/pdf/2510.24701
DeepSeek-V3.1~\cite{liu2024deepseekv3}             & 29.8 & 30.0 & 49.2 & 63.1 & 71.0 & 61.2 & 83.7 &  --  \\
% Results of DeepSeek-V3.1-Terminus from: https://huggingface.co/deepseek-ai/DeepSeek-V3.1-Terminus
% DeepSeek-V3.1-Terminus~\cite{liu2024deepseekv3}  & --   & 38.5 & 45.0 & --   & --   & --   & --   &      \\
% Results of DeepSeek-V3.2 from: https://huggingface.co/MiniMaxAI/MiniMax-M2
% and from: https://huggingface.co/moonshotai/Kimi-K2-Thinking
DeepSeek-V3.2~\cite{liu2024deepseekv3}             & 27.2 & 40.1 & 47.9 & 63.5 & 71.0 & --   & 80.2 & 38.5 \\
% Results of Kimi-K2-0905 from: https://huggingface.co/moonshotai/Kimi-K2-Thinking
% and from: https://huggingface.co/MiniMaxAI/MiniMax-M2
Kimi-K2-0905~\cite{team2025kimik2}                 & 21.7 & 7.4  & 22.2 & 60.2 & 61.0 & --   & 58.1 & 25.2 \\
% Results of Claude-4-Sonnet from: https://arxiv.org/pdf/2510.24701
Claude-4-Sonnet~\cite{anthropic2025claude4_5}      & 20.3 & 12.2 & 29.1 & 68.3 & 64.6 & 61.7 & 80.7 & --   \\
% Results of Claude-4.5-Sonnet from: https://huggingface.co/MiniMaxAI/MiniMax-M2
Claude-4.5-Sonnet~\cite{anthropic2025claude4_5}    & 24.5 & 19.6 & 40.8 & 71.2 & 66.0 & --   & 85.0 & 53.4 \\
% Results of OpenAI-o3 from: https://arxiv.org/pdf/2510.24701
OpenAI-o3~\cite{openai2025o3_o4mini}               & 24.9 & 49.7 & 58.1 & --   & 67.0 & 71.7 & 84.0 & 17.1 \\
% Results of OpenAI-o4-mini from: https://arxiv.org/pdf/2510.24701
OpenAI-GPT-5-high~\cite{openai2025gpt5}                   & 35.2 & 54.9 & 65.0 & 76.4 & 77.8 & --   & --   & 51.4 \\
\midrule
\multicolumn{8}{l}{\textit{Research Agents}} \\
\midrule
% Gemini DeepResearch                              & 26.9 & --   & --   & --   & --   & --   & --   &      \\
OpenAI DeepResearch~\cite{openai2025deepresearch}  & 26.6 & 51.5 & 42.9 & 67.4 & --   & --   & --   & --   \\
ChatGPT-Agent~\cite{openai2025chatgpt_agent}       & 41.6 & 68.9 & --   & --   & --   & --   & --   & --   \\
Kimi-Researcher~\cite{moonshot2025kimi_researcher} & 26.9 & --   & --   & --   & 69.0 & --   & 78.8 & 36.0 \\
WebExplorer-8B-RL~\cite{liu2025webexplorer}        & 17.3 & 15.7 & 32.0 & 50.0 & 53.7 & 62.7 & 75.7 & --   \\
DeepMiner-32B-RL~\cite{tang2025beyond}             & --   & 33.5 & 40.1 & 58.7 & 62.0 & --   & --   & --   \\
AFM-32B-RL~\cite{li2025afm}                        & 18.0 & 11.1 & --   & 55.3 & --   & 63.0 & --   & --   \\
SFR-DeepResearch-20B~\cite{nguyen2025sfr}          & 28.7 & --   & --   & 66.0 & --   & --   & 82.8 & --   \\
Tongyi-DeepResearch-30B~\cite{team2025tongyi}      & 32.9 & 43.4 & 46.7 & 70.9 & 75.0 & 72.2 & 90.6 & --   \\
\midrule
\rowcolor{rowhl} MiroThinker-v1.0-8B               & 21.5{\scriptsize$\pm 0.4$} & 31.1{\scriptsize$\pm 1.6$} & 40.2{\scriptsize$\pm 2.9$} & 66.4{\scriptsize$\pm 3.2$} & 60.6{\scriptsize$\pm 3.8$} & 60.6{\scriptsize$\pm 0.8$} & 80.6{\scriptsize$\pm 0.5$} & 40.4{\scriptsize$\pm 2.6$} \\
\rowcolor{rowhl} MiroThinker-v1.0-30B              & 33.4{\scriptsize$\pm 0.2$} & 41.2{\scriptsize$\pm 1.3$} & 47.8{\scriptsize$\pm 1.1$} & 73.5{\scriptsize$\pm 2.6$} & 70.6{\scriptsize$\pm 2.2$} & 61.0{\scriptsize$\pm 0.2$} & 85.4{\scriptsize$\pm 0.8$} & 46.8{\scriptsize$\pm 3.2$} \\
\rowcolor{rowhl} MiroThinker-v1.0-72B              & 37.7{\scriptsize$\pm 0.5$} & 47.1{\scriptsize$\pm 0.7$} & 55.6{\scriptsize$\pm 1.1$} & 81.9{\scriptsize$\pm 1.5$} & 77.8{\scriptsize$\pm 2.6$} & 62.1{\scriptsize$\pm 0.6$} & 87.1{\scriptsize$\pm 0.9$} & 51.0{\scriptsize$\pm 2.0$} \\
\bottomrule
\end{tabularx}
\end{table}

\section{Experiments}
\subsection{Experimental Setup}

\paragraph{Evaluation Benchmarks.}  
We use the Qwen2.5 and Qwen3 models~\cite{yang2025qwen3} as initialization.
After training, we evaluate MiroThinker v1.0 models across a diverse suite of agentic benchmarks: Humanity’s Last Exam (HLE) \citep{phan2025hle}; BrowseComp \citep{wei2025browsecomp} and BrowseComp-ZH \citep{zhou2025browsecomp_zh}; GAIA \citep{mialon2023gaia}; xBench-DeepSearch \citep{chen2025xbench}; WebWalkerQA \citep{wu2025webwalker}; FRAMES \citep{krishna2025fact}; and SEAL-0~\cite{pham2025sealqa}.
To ensure fair comparison with prior works, we follow the standard evaluation protocol and report results on the 2,158 text-only subset of Humanity’s Last Exam and the 103 text-only subset of GAIA. For other benchmarks, we report the model’s performance on the full test set.
Note, to prevent potential information leakage (\emph{e.g.}, searching benchmark answers from HuggingFace), access to HuggingFace has been explicitly disabled in these tools.

\paragraph{Evaluation Protocol.}
We report all benchmark results using a simple ReAct-style agent to fully demonstrate the strength of our MiroThinkers.
We adopt fixed inference settings to guarantee stability and reproducibility: temperature = 1.0, top-p = 0.95, maximum turns = 600, context length = 256K tokens, and maximum output length = 16,384 tokens. 
For context management, the retention budget is set to 5.
For benchmarks exhibiting high per-question variance, we conduct $k$ independent runs and report the averaged score, denoted as avg@k. The specific $k$ values for each benchmark are detailed in the caption of Table~\ref{tab:benchmarks-fast}. All benchmark performances are evaluated using LLM-as-a-Judge.
Specifically, GAIA, WebWalkerQA, xBench-DeepSearch, BrowseComp, and BrowseComp-ZH are judged with gpt-4.1-2025-04-14, while Humanity’s Last Exam follows its official setup using o3-mini-2025-01-31.

\begin{figure}[t]
  \centering
  \begin{subfigure}{0.24\linewidth}
    \includegraphics[width=\linewidth]{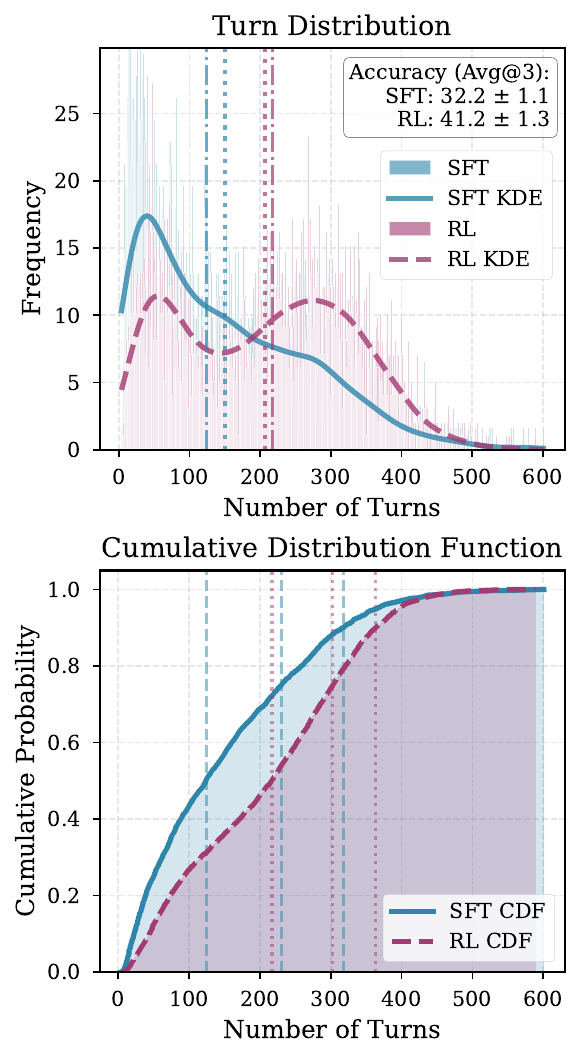}
    \caption{BrowseComp}
    \label{fig:isl-a}
  \end{subfigure}
  \hfill
  \begin{subfigure}{0.24\linewidth}
    \includegraphics[width=\linewidth]{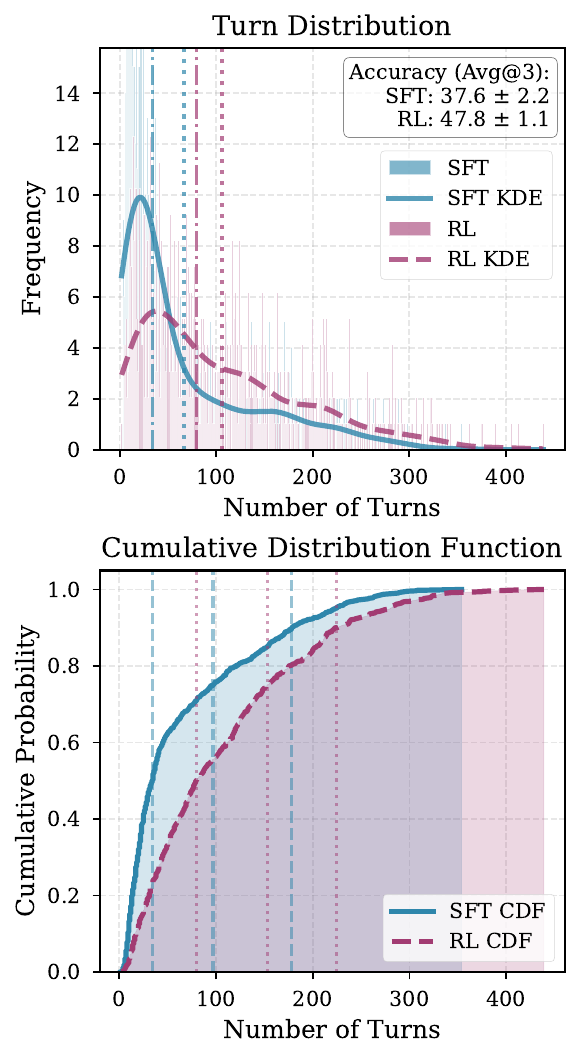}
    \caption{BrowseComp-ZH}
    \label{fig:isl-b}
  \end{subfigure}
  \hfill
  \begin{subfigure}{0.243\linewidth}
    \includegraphics[width=\linewidth]{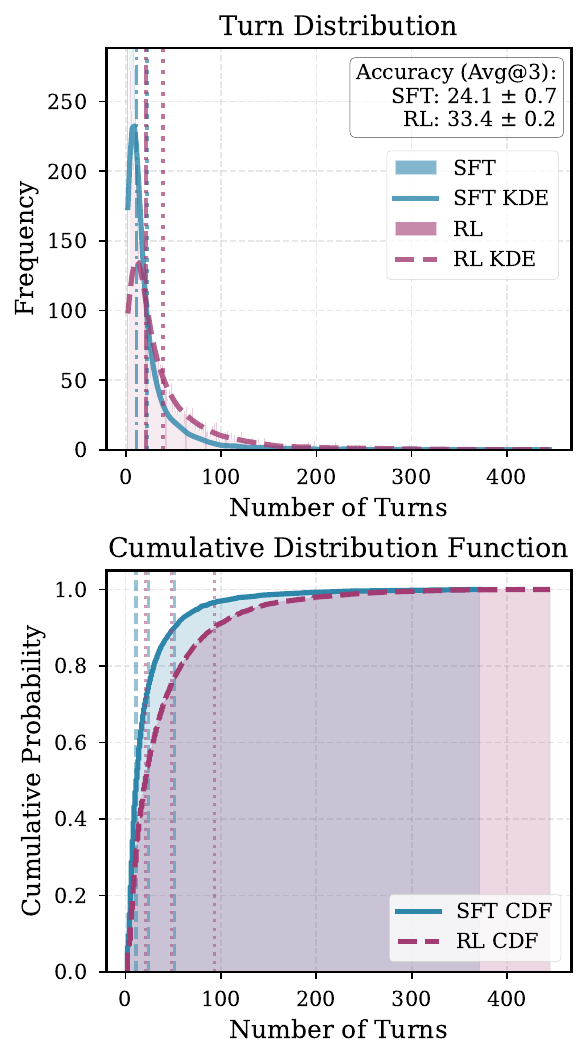}
    \caption{HLE}
    \label{fig:isl-c}
  \end{subfigure}
  \hfill
  \begin{subfigure}{0.24\linewidth}
    \includegraphics[width=\linewidth]{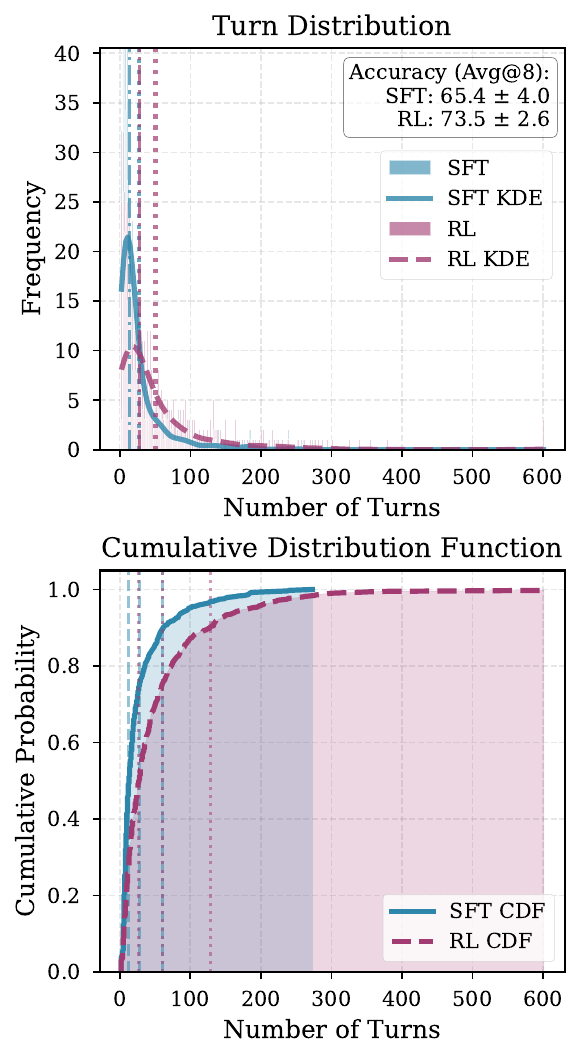}
    \caption{GAIA}
    \label{fig:isl-d}
  \end{subfigure}

  \caption{Illustration of interactive scaling. Reinforcement learning training leads to a substantial increase in the number and depth of agent–environment interactions, resulting in consistently improved task performance across benchmarks. All results are from MiroThinker-v1.0-30B.}
  \label{fig:interactive-scaling-law}
\end{figure}

\subsection{Overall Performance}
MiroThinker establishes a new state-of-the-art on the GAIA benchmark~\cite{mialon2023gaia} with a score of 81.9\%, surpassing the previous leading model, MiniMax-M2~\cite{minimax2025m2}, by 6.2 percentage points. 
The model's capabilities are further demonstrated on the extremely challenging Humanity's Last Exam~\cite{phan2025hle}, where it achieves a score of 37.7\%. 
This result outperforms the proprietary state-of-the-art model, GPT-5-high~\cite{openai2025gpt5}, by 2.5 percentage points while utilizing the identical Python and search toolset.

Furthermore, MiroThinker delivers highly competitive performance on the BrowseComp~\cite{wei2025browsecomp} and SEAL-0~\cite{pham2025sealqa} benchmarks, with scores of 47.1\% and 51.0\% respectively, placing it on par with advanced proprietary systems such as OpenAI DeepResearch~\cite{openai2025deepresearch}, OpenAI o3~\cite{openai2025o3_o4mini}, and Anthropic Claude 4.5~\cite{anthropic2025claude4_5}. 
In the domain of multilingual research, MiroThinker shows exceptional mastery by setting new open-source records on Chinese benchmarks, achieving 55.6\% on BrowseComp-ZH~\cite{zhou2025browsecomp_zh} and 77.8\% on xbench-DeepSearch~\cite{chen2025xbench}.

Our commitment to excellence extends across all scales. 
The 8B and 30B variants of MiroThinker also achieve state-of-the-art performance within their respective size classes, providing the community with access to powerful deep research models at various model scales.

\subsection{Interactive Scaling}
We examine how RL reshapes agent–environment interaction patterns.
As shown in Figure~\ref{fig:interactive-scaling-law}, the RL-tuned MiroThinker-v1.0-30B model exhibits substantially longer and deeper interaction trajectories than its SFT counterpart across BrowseComp~\cite{wei2025browsecomp}, BrowseComp-ZH~\cite{zhou2025browsecomp_zh}, HLE~\cite{phan2025hle}, and GAIA~\cite{mialon2023gaia}.
Guided by verifiable rewards, RL enables the model to explore more exhaustive solution paths with significantly greater interaction depth, systematically probing multiple strategies and validating intermediate results before reaching conclusions.

This behavioral shift directly correlates with improved accuracy, yielding 8–10 point gains on average.
We refer to this consistent relationship between interaction depth and performance as the interactive scaling: as the frequency and depth of tool-augmented interactions increase, research reasoning capability improves correspondingly.
This forms the third dimension of scaling, alongside model size and context length, defining MiroThinker’s pathway toward more general agentic intelligence.

\subsection{Limitations}

We have identified several limitations in the current version of the model, which we plan to address in future updates. These limitations include the following:

\paragraph{Tool-Use Quality under Interactive Scaling}
Interactive scaling enables richer and more complex tool interactions, but also exposes limitations in tool-use quality. We find that the RL-tuned model invokes external tools more frequently than the SFT model, yet a portion of these invocations yield marginal or redundant contributions. This indicates that scaling improves agentic performance, but further optimization is needed to enhance tool-use efficiency and action quality.

\paragraph{Overlong Chain-of-Thought} 
Reinforcement learning tends to encourage the model to produce longer responses to improve accuracy, which can result in excessively long, repetitive, and less readable reasoning chains. This, in turn, slows down task completion and degrades user experience.

\paragraph{Language Mixing} 
For non-English inputs, the model’s responses may exhibit multilingual mixing. For instance, when a user query is in Chinese, the model’s internal reasoning or intermediate outputs may contain a blend of English and Chinese elements, which may lead to suboptimal performance in Chinese.

\paragraph{Limited Sandbox Capability} 
The model is not yet fully proficient in utilizing code-execution and file-management tools. It may occasionally generate code or commands that lead to sandbox timeouts, or misuse the code-execution tool to read web pages or PDFs, tasks that would be far more efficiently handled by dedicated web-scraping tools. Moreover, the model sometimes demonstrates insufficient familiarity with sandbox ID management, frequently forgetting to initialize a sandbox before invoking related operations.
\section{Conclusions}

We introduce MiroThinker v1.0, an open-source research agent that advances tool-augmented reasoning through model, context, and interactive scaling. By extending scaling into the interaction dimension, MiroThinker shows that research capability improves not only with larger models or longer context, but also with deeper and more frequent agent–environment interactions that enable error correction and knowledge acquisition. Our experiments demonstrate predictable gains from interactive scaling across diverse benchmarks, establishing interaction depth as a third critical axis for building next-generation research agents. We hope MiroThinker provides a strong baseline and an open platform for further exploration of interaction-scaled, agentic intelligence.
% \clearpage
\bibliography{main}

\newpage

\section*{A \quad Contributions}

The listing of authors is in alphabetical order based on their last names. 

\begin{multicols}{3}
\begin{spacing}{1.1}

MiroMind Team \\
Song Bai \\
Lidong Bing \\
Carson Chen \\
Guanzheng Chen \\
Yuntao Chen \\
Zhe Chen \\
Ziyi Chen \\
Jifeng Dai \\
Xuan Dong \\
Wenhan Dou \\
Yue Deng \\
Yunjie Fu \\
Junqi Ge \\
Chenxia Han \\
Tammy Huang \\
Zhenhang Huang \\
Jerry Jiao \\
Shilei Jiang \\
Tianyu Jiao \\
Xiaoqi Jian \\
Lei Lei \\
Ruilin Li \\
Gen Luo \\
Tiantong Li \\
Xiang Lin \\
Ziyuan Liu \\
Zhiqi Li \\
Jie Ni \\
Qiang Ren \\
Pax Sun \\
Shiqian Su \\
Chenxin Tao \\
Bin Wang \\
Wenhai Wang \\
Haonan Wang \\
James Wang \\
Jin Wang \\
Jojo Wang \\
Letian Wang \\
Shizun Wang \\
Weizhi Wang \\
Zixuan Wang \\
Jinfan Xu \\
Sen Xing \\
Chenyu Yang \\
Hai Ye \\
Jiaheng Yu \\
Yue Yu \\
Muyan Zhong \\
Tianchen Zhao \\
Xizhou Zhu \\
Yanpeng Zhou \\
Yifan Zhang \\
Zhi Zhu \\

\end{spacing}
\end{multicols}

\end{document}